\documentclass[10pt,twocolumn,letterpaper]{article}

\usepackage[pagenumbers]{cvpr} 

\usepackage{graphicx}
\usepackage{booktabs}
\usepackage{amsmath}
\usepackage{amssymb}
\usepackage{tablefootnote}
\usepackage{multicol}
\newcommand{\mathbit}[1]{#1}
\usepackage[pagebackref,breaklinks,colorlinks]{hyperref}

\usepackage[capitalize]{cleveref}
\crefname{section}{Sec.}{Secs.}
\Crefname{section}{Section}{Sections}
\Crefname{table}{Table}{Tables}
\crefname{table}{Tab.}{Tabs.}

\def\cvprPaperID{5536} 
\def\confName{CVPR}
\def\confYear{2023}

\begin{document}

\title{Focal-PETR: Embracing Foreground for Efficient Multi-Camera 3D Object Detection}

\author{Shihao Wang\thanks{equal contribution}\\{\tt\small wangshihao@bit.edu.cn}
\and
Xiaohui Jiang\footnotemark[1]\\{\tt\small xhjiang@bit.edu.cn}
\and
Ying Li\thanks{Corresponding author}\\{\tt\small ying.li@bit.edu.cn}
\and
Beijing Institute of Technology\\
}

\maketitle

\begin{abstract}
The dominant multi-camera 3D detection paradigm is based on explicit 3D feature construction, which requires complicated indexing of local image-view features via 3D-to-2D projection. Other methods implicitly introduce geometric positional encoding and perform global attention (e.g., PETR) to build the relationship between image tokens and 3D objects. The 3D-to-2D perspective inconsistency and global attention lead to a weak correlation between foreground tokens and queries, resulting in slow convergence. We propose \textbf{Focal-PETR} with instance-guided supervision and spatial alignment module to adaptively focus object queries on discriminative foreground regions. \textbf{Focal-PETR} additionally introduces a down-sampling strategy to reduce the consumption of global attention. Due to the highly parallelized implementation and down-sampling strategy, our model, without depth supervision, achieves leading performance on the large-scale nuScenes benchmark and a superior speed of \textbf{30 FPS} on a single RTX3090 GPU. Extensive experiments show that our method outperforms PETR while consuming \textbf{3x} fewer training hours. The code will be made publicly available.
\end{abstract}

\vspace{-0.5cm}
\section{Introduction}
\label{sec:intro}

Camera-based 3D object detection, compared with LiDAR-based counterparts has attracted immense attention in recent years due to its low cost for deployment and dense semantic information \cite{ma2022vision, ma20223d}. Multiple cameras with overlapping regions usually need to be setup for camera-based panoramic perception. To efficiently fuse information from overlapping regions, a unified representation for arbitrary camera rigs \cite{huang2021bevdet, li2022bevformer, zhou2022cross, wang2022detr3d, li2022hdmapnet} has been widely concerned.

\begin{figure}[t]
\centering
\includegraphics[scale=0.38]{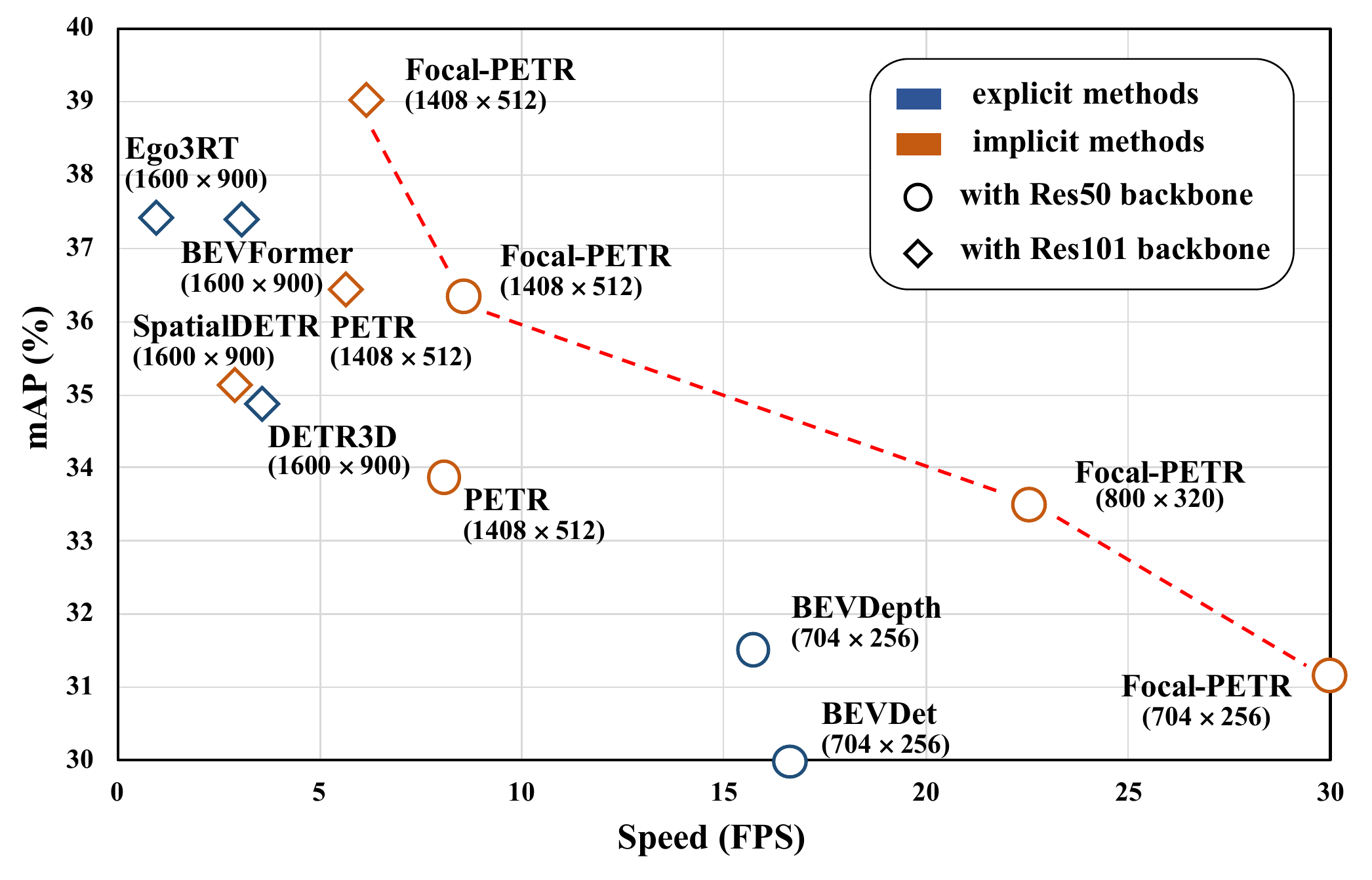}
\vspace{-0.4cm}
\caption{Speed-accuracy trade-off of different models with a single time stamp input on nuScenes val set. The inference speed is calculated on a single RTX3090 GPU. Due to our focal sampling and high parallelism, we have achieved superior performance. The corresponding input resolutions are in brackets. More detailed analysis can be found in Table \ref{val_tab}.}
\vspace{-0.5cm} 
\label{speed_vs_acc}
\end{figure}

To explore a unified 3D representation, the dominant multi-camera 3D detection models are based on explicit 3D feature construction, which mainly relies on precise prediction of depth distribution or perspective projection\cite{philion2020lift, li2022bevformer,  wang2022detr3d, li2022bevdepth}. These explicit models have achieved leading performance on benchmarks, while they are unfriendly for parallel computing on GPU devices due to complicated feature indexing operations. Another paradigm has been proposed by simply converting 2D image features into 3D position awareness through implicit position embedding. Such paradigm has achieved comparable performance with advantages of global modeling capabilities and parallel computing performance, as illustrated in Figure \ref{speed_vs_acc}. While conciseness and effectiveness, the long training schedule and huge memory consumption limit its large-scale application in autonomous driving compared with the explicit paradigm. We suggest that the long training schedule is largely on account of the misalignment between the training objectives and features gathered by object queries. The explicit methods directly align features through auxiliary depth supervision or accurate 3D-to-2D projection. However, this mechanism in implicit methods has not been well explored.

\begin{figure}[t]
\centering
\includegraphics[scale=0.12]{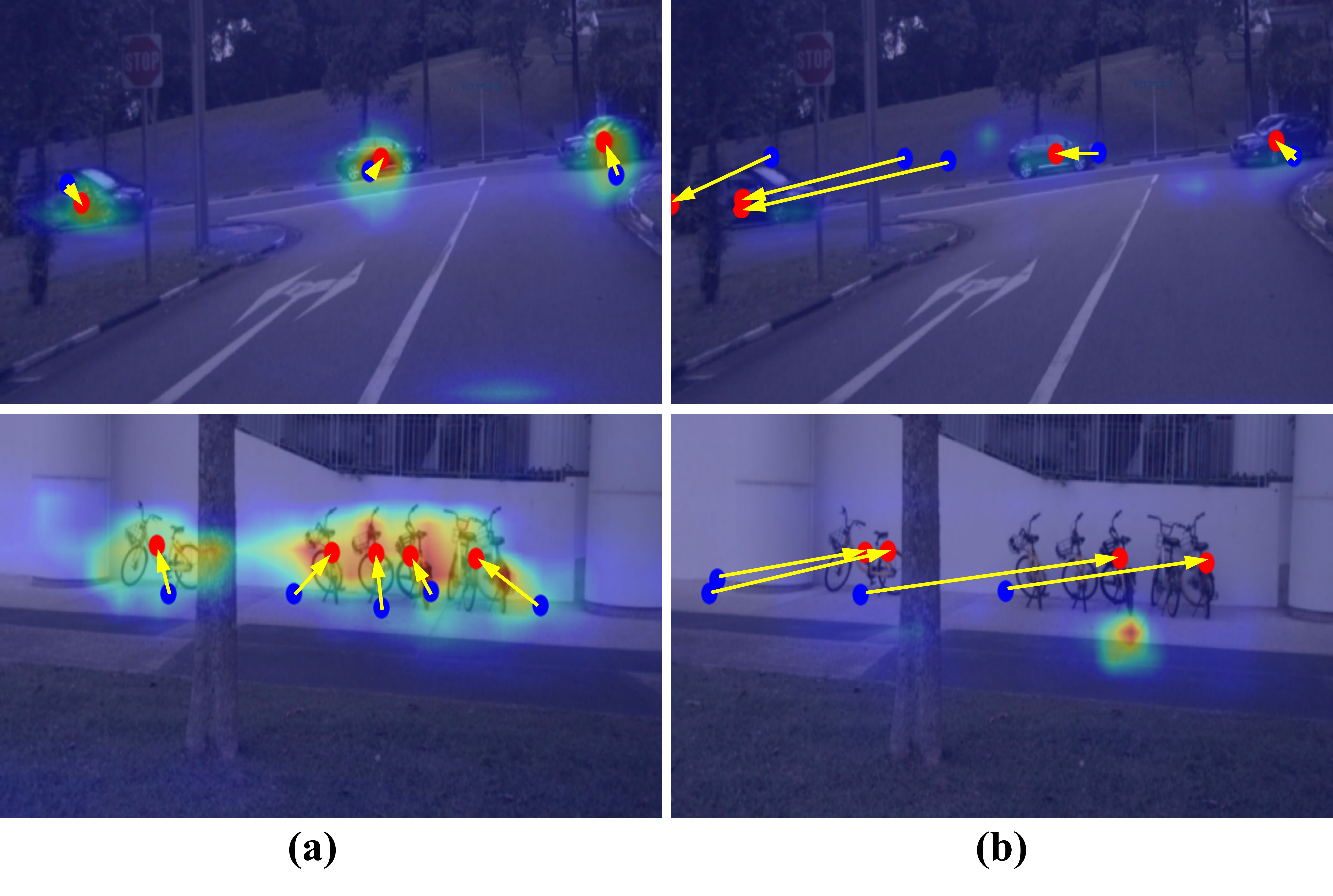}
\caption{Comparison of attention weight maps for (a) Focal-PETR and (b) PETR with 24 training epochs without CBGS \cite{zhu2019class}. The reference points corresponding to different layers are projected onto the image plane. The blue and red dots represents initialized anchor points and final predictions respectively. It can be seen that the attention map of our Focal-PETR focuses on the foreground objects.}
\vspace{-0.5cm}
\label{fig2}
\end{figure}

 We explore PETR\cite{liu2022petr, liu2022petrv2}, which performs best among existing implicit methods, and suggest two limitations of this paradigm: (i) semantic ambiguity and (ii) spatial misalignment. The reason for (i) is the similar content embedding of image tokens. As shown in Figure \ref{fig2}, the object queries give a nearly uniform attention weights on non-local regions and their initialized reference points are far from the predictions. We conclude that the lack of discrimination in foreground features makes queries difficult to focus on extremities. After subsequent attention layers, the obtained object queries for positive sample matching are not closely correlated with relevant foreground features during the training phase. The reason for (ii) is that the geometric cues only play the role of positional embedding and are not considered as a part of content collected by object queries (see Figure \ref{fig4}). Thus, the search mode of object queries is semantic-biased in the cross-attention modules, which means that the calculation of query-to-feature similarity is spatial insensitive. 
 
We propose Focal-PETR, a semantic-aggregated and spatial-aligned framework to strengthen the discrimination and spatial sensitivity of foreground tokens. Specifically, three auxiliary tasks, namely class-aware, IoU-aware and centroid-aware strategies, are adopted to explore salient tokens in an instance-guided manner. Based on them we verify that the view transformer process can be considerably achieved by condensing the foreground tokens into a smaller set. This design greatly reduces the computing consumption of global cross attention. Additionally, a spatial alignment module is designed to embed the spatial information into image tokens, which enables the decoder to effectively search the content.

Figure \ref{speed_vs_acc} shows that the proposed Focal-PETR preserves and enhances the advantages of implicit 3D detection paradigm, which can be highly parallel implemented. On nuScenes \cite{caesar2020nuscenes} val set, the lightweight model with Res50 backbone achieves superior accuracy and speed trade-off (i.e., 31.1\% mAP and 30.0 FPS on a single RTX3090). Extensive experiments and analysis on Section \ref{sec:experiment} justify the effectiveness and efficiency of the method.

To summarize, the contributions of this paper are:
\begin{itemize}

    \item We first identify the semantic ambiguity and spatial misalignment in existing implicit 3D detection paradigms, causing suboptimal discriminant features extraction and slow convergence speed.

    \item We propose Focal-PETR by introducing focal sampling and spatial alignment modules. Our method mitigates the aforementioned problems and efficiently focuses on the foreground tokens. We also analyze the computational consumption and memory footprint to further validate the efficiency of our proposed method.

    \item Experiments on the large-scale nuScenes benchmark show superior efficiency and state-of-the-art performance (46.5\% mAP and 51.6\% NDS) of the proposed Focal-PETR with a single time stamp input.
\end{itemize}

\begin{figure*}[t]
    \centering
    \includegraphics[scale=0.50]{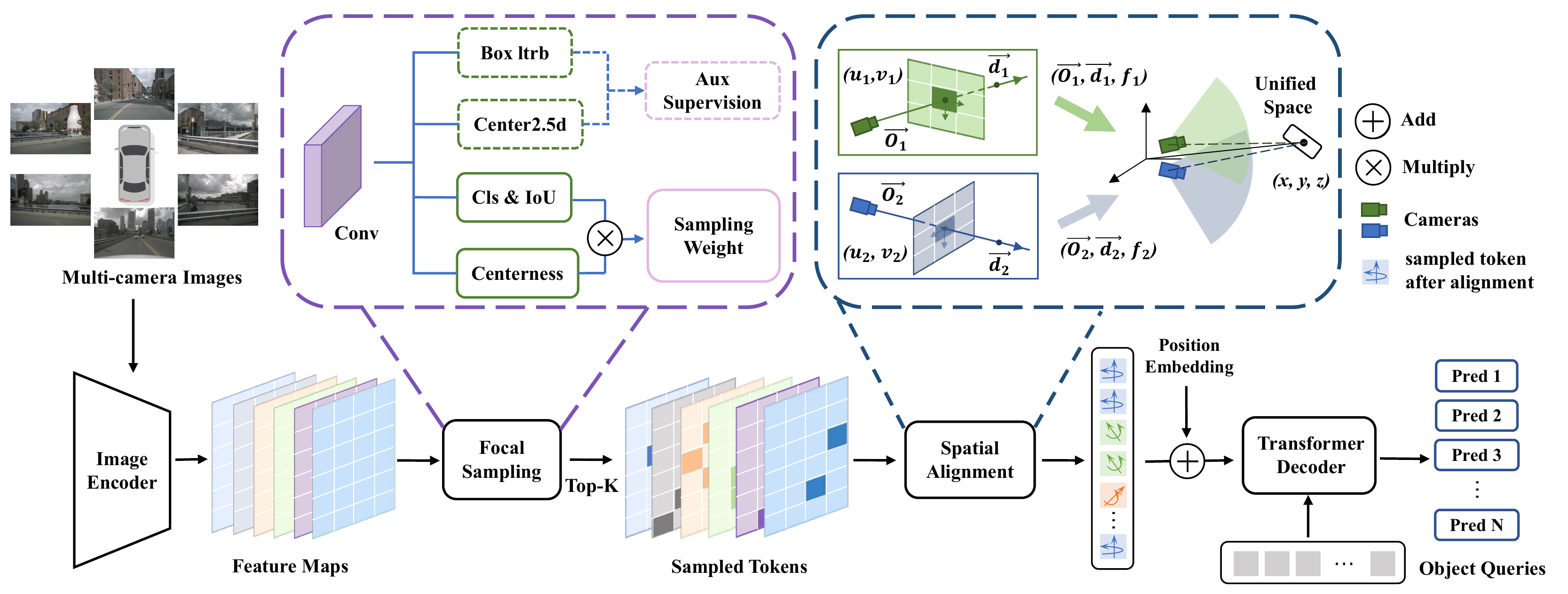}
    \caption{Overview of Focal-PETR. The multi-camera images are first fed into an image encoder to learn a high-level 2D representation. Then a focal sampling module is applied to distinguish discriminative foreground tokens. The auxiliary supervision which is turned off at inference is shown in dashed lines. The sampled tokens are further fed into the spatial alignment module to convert 2D features into a unified 3D space, according to parametric conical frustum. Finally, Transformer decoder is employed to generate 3D bounding boxes. $ \boldsymbol{\mathbit\vec{o}} $ and $ \boldsymbol{\mathbit\vec{d}} $ represent the optical center and direction of a specific pixel ray. $ \boldsymbol{\mathbit{f}} $ is the camera intrinsic, additionally employed to describe frustum cone viewed by each ray.}
    \vspace{-0.5cm}
    \label{fig3}
\end{figure*}

\section{Related Work}
\label{sec:related work}

\subsection{Multi-camera 3D Object Detection}
3D object detection based on unified representation has attracted increasing attention for multi-sensors fusion. Previous works such as Pseudo-LiDAR \cite{wang2019pseudo}, OFT \cite{roddick2018orthographic}, LSS \cite{philion2020lift}, and BEVDet \cite{huang2021bevdet} explicitly predict the depth distribution to lift 2D features to 3D space. DETR3D \cite{wang2022detr3d} first projects the predefined queries onto images and then adopts attention mechanism to model the relation with the multi-view features respectively. BEVFomrer \cite{li2022bevformer} further extends this idea by using dense queries and temporal fusion. These methods explicitly index the local image feature from 2D perspective-view to 3D space, facilitating the alignment between training targets and image features. Other works \cite{zhou2022cross, liu2022petr, liu2022petrv2, dollspatialdetr} model the view transformation by implicitly encoding geometric information, building the interaction between 3D queries and image tokens. Due to the powerful modeling capabilities of Transformer \cite{vaswani2017attention}, models based on implicit positional encoding can extract global information in a parallel way but suffer from slow convergence and huge memory complexity. We analyze the mechanism of implicit paradigm and suggest that the weak ability to locate foreground information impairs the representation of object queries, resulting in the misalignment of training targets and semantic content. We perform auxiliary tasks to adaptively focus the attention on salient regions.

\subsection{Fast Converge of Query-based detectors}
Query-based detectors have been widely studied due to their high performance and simplicity, while the slow training convergence limits the large-scale deployment \cite{li2022dn}. Some works \cite{li2022dn, dai2021dynamic, gao2022adamixer, wang2021anchor, liu2022dab, sun2021rethinking, meng2021conditional, zhu2020deformable} have tried to solve this problem. Several of them attempt to improve the network structure by taking local attention operation, such as Deformerable DETR \cite{zhu2020deformable}, Dynamic DETR \cite{dai2021dynamic}, AdaMixer \cite{gao2022adamixer}. Besides, initializing queries with meaningful information has been investigated. Anchor DETR \cite{wang2021anchor} and DAB-DETR \cite{liu2022dab} interpret query as 2D reference points or 4D anchor boxes. Conditional DETR \cite{meng2021conditional} combines the content and position information together to focus every query on a specific spatial space. DN-DETR \cite{li2022dn} considers the influence of instable bipartite graph matching and introduces query denoising to mitigate slow convergence.  
The training objectives and features of 2D query-based detectors are all located in the same perspective view. While the 3D-to-2D query-based detectors are more difficult to converge due to the weak correspondence between object queries and the gathered features. We propose a spatial alignment module to enhance the spatial sensitivity of object queries on the basis of semantic correspondence.

\subsection{Ranking and Sampling in Object Detection}
Convolution based dense detectors \cite{tian2019fcos, lin2017focal, ren2015faster} with one-to-many label assignment criteria have led the mainstream of object detection. Due to the non-uniformity and sparsity of foreground information in visual signal, duplicate predictions of objects are inevitable. Generally, additional NMS \cite{ren2015faster} and bounding box quality ranking strategies \cite{wu2020iou, li2020generalized}, such as classification, centerness \cite{tian2019fcos} and IoU scores, are introduced to eliminate the redundant predictions. DETR \cite{carion2020end} outperforms competitive Faster R-CNN \cite{ren2015faster} baseline by relying on the Transformer \cite{vaswani2017attention} architecture and one-to-one assignment strategy \cite{carion2020end}. However, the huge computation and memory consumption of attention in Transformer limit its further development. Recent works conduct sparse sampling for attention mechanism \cite{zhu2020deformable, wang2021pnp, roh2021sparse}. Deformable DETR
\cite{zhu2020deformable} performs learnable adjacent sampling instead of the entire image features. PnP-DETR \cite{wang2021pnp} and Sparse DETR \cite{roh2021sparse} sample a fraction of salient tokens in an unsupervised way for the subsequent encoder and decoder. In this paper, we propose focal sampling strategy based on the instance-guided supervision. Simply sampling foreground features via detection quality ranking achieves competitive performance and verify the training consistency between 2D and 3D object detection tasks.

\section{Method}
\label{sec:mothod}

Existing 3D detection models based on implicit positional encoding \cite{dollspatialdetr, liu2022petr, liu2022petrv2} directly utilize pixel-level features extracted by an image encoder as the minimum unit to process view transformation. We suggest that the direct utilization of pixel representation makes object queries hard to focus on foreground features. Thus, we attempt to interpret the features as discriminative instances. In the following, we first revisit the detection pipeline proposed in PETR \cite{liu2022petr}, which is high parallelism and suitable for fusing heterogeneous features between instances. Then we elaborate on our instance-guided down-sampling strategy and {spatial alignment module}.

\subsection{Preliminaries: PETR}
PETR is built upon the Transformer decoder architecture \cite{vaswani2017attention}. Its core components include image encoder, positional encoder and detection head. Combining sparse BEV queries with cross-attention mechanism, PETR refines the detection prediction using implicit features enhanced by 3D positional embedding in a cascade manner. A brief review of the PETR pipeline is made as follows.

\textbf{3D Positional Encoder.} Given an image $I_j\in\mathbb{R}^{3\times H\times W}$ captured from one of N $(j\in\{1,2,..., N\})$ surround cameras with its corresponding intrinsic matrix $K_j\in\mathbb{R}^{3\times3}$ and extrinsic matrix $P_j\in\mathbb{R}^{4\times4}$, a tracing ray corresponding to each pixel center in a unified coordinate system $\boldsymbol{\mathbit{r}}_j^{u,v}\left(t\right)\epsilon\mathbb{R}^3=\left\{\boldsymbol{\mathbit{o}}_\mathbit{j}+t\boldsymbol{\mathbit{d}}_j^{u,v}\middle| t\epsilon\mathbb{R},\boldsymbol{\mathbit{o}}\epsilon\mathbb{R}^3,\boldsymbol{\mathbit{d}}\epsilon\mathbb{R}^3\right\}$ can be derived. Each pixel position $({\mathbit{u},\mathbit{v}})$ on the specific camera emits a unique ray along the direction $\boldsymbol{d}_j^{u,v}$, which passes through the optical center $\boldsymbol{\mathbit{o}}_\mathbit{j}$ of the corresponding camera. Based on the aforementioned ray equation, the linear-increasing discretization (LID) \cite{tang2020center3d} is adopted to approximately sample the rays at different distances $t_\mathbit{i}$, where $\mathbit{i}$ is the depth bin index in LID. Then the coordinates $\boldsymbol{\mathbit{C}}_j^{u,v}$ of sampled points are normalized and fed into a 2-layer multi-layer perception (MLP) $\varphi$. The resulting positional embedding is noted as $\boldsymbol{\mathbit{E}}_j^{u,v}$. This process can be abstracted as: 
\begin{equation}
    \boldsymbol{\mathbit{C}}_j^{u,v}=Concat[\boldsymbol{\mathbit{r}}_j^{u,v}\left(t_1\right),\ \boldsymbol{\mathbit{r}}_j^{u,v}\left(t_2\right),\ldots,\boldsymbol{\mathbit{r}}_j^{u,v}\left(t_i \right)]
\end{equation}
\begin{equation}
\boldsymbol{\mathbit{E}}_j^{u,v}= \varphi(Norm(\mathit{\mathbf{C}}_j^{u,v}))
\end{equation}

\textbf{Detection Head.} The detection head is composed of conventional Transformer decoder layers \cite{vaswani2017attention}, which interacts queries $\boldsymbol{q}_{L}$ of layer \textit{L} with 3D positional enhanced features  $(\boldsymbol{k})$ from image encoder. To enable the features $\boldsymbol{\mathbit{F}}_j^{u,\ v}$ aggregated by queries be stably supervised, the queries are defined as a group of learnable spatial anchor points. The interaction mentioned above can be expressed as:

\begin{equation}
    \label{eq3}
    [\boldsymbol{\mathbit{k}},\boldsymbol{\mathbit{v}}]=[\boldsymbol{\mathbit{F}}_j^{u,\ v}+\boldsymbol{\mathbit{E}}_j^{u,v},\ \boldsymbol{\mathbit{F}}_j^{u,\ v}]
\end{equation}
\begin{equation}
    \boldsymbol{\mathbit{q}}_\mathbit{L}=Softmax(\boldsymbol{\mathbit{q}}_{\mathbit{L}-\mathbf{1}}\cdot{\ \boldsymbol{\mathbit{k}}}^T)\cdot\boldsymbol{\mathbit{v}}
\end{equation}

For simplicity, we ignore the scaling factor. It should be noticed that, the 3D implicit features are only used as keys $(\boldsymbol{k})$ to interact with those queries. It is to say that the values $(\boldsymbol{v})$ used for weighted sum are geometry-ignored. Finally, queries after multi-layer refinement are sent to finish classification and regression tasks respectively. For more details, please refer to the original paper of PETR \cite{liu2022petr}.

\begin{figure}[t]
    \centering
    \includegraphics[scale=0.52]{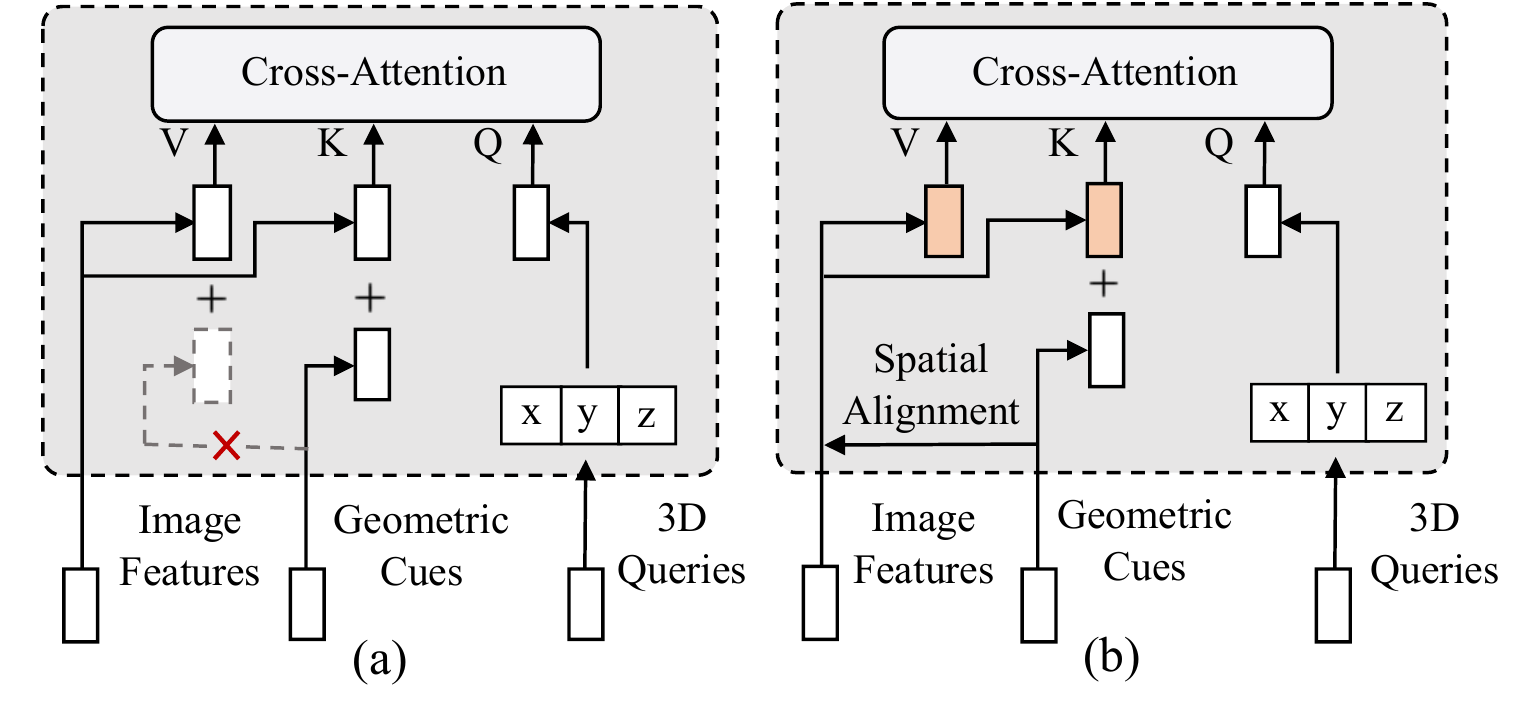}
    \vspace{-0.2cm}
    \caption{Content composition of query, key and value. (a) structure in PETR; (b) structure in Focal-PETR. The rectangle boxes in orange represent features after spatial alignment.}
    \vspace{-0.5cm}
    \label{fig4}
\end{figure}

\subsection{Focal Sampling Strategy}

The purpose of our focal sampling is to distinct features in instance-level and distill the representative tokens while ensuring high recall. Specifically, we first adopt an instance-guided way and decouple the foreground information into three categories, including semantic discriminability, object integrity and position sensitivity. To select corresponding features respectively, three sub-modules are conducted, which are class-aware, IoU-aware and centroid-aware modules. Several convolutional heads are appended to the image encoder to predict the quality score of attributes (see Figure \ref{fig2}). Focal sampling module is lightweight and has negligible impact on the inference time. The above three down-sampling strategies mainly emphasize the importance of positive samples assignment and loss definition.

\textbf{Class-aware Sampling.} In order to exploit the discriminative foreground information as soon as possible, additional 2D object detection task to infer the objectiveness is performed. Specifically, we follow FCOS \cite{tian2019fcos} to supervise the classification score $c$ and the normalized distances $(l^{*},t^{*},r^{*},b^{*})$ from the location to the four sides of the bounding box. Beseides, Hugarian matching [2] that considers both classification and location costs is adopted for positive sample selection, which can supervise the network to generate high-recall predictions without post-processing \cite{wang2021end, sun2021makes}. For simplicity, Focal loss \cite{lin2017focal} and L1 loss are used for classification and regression supervision, respectively.

\textbf{IoU-aware Sampling.} Previous works \cite{lian2022monojsg} have indicated that the precise estimation of 2D pose attributes is highly related to 3D geometric information. Therefore, only sampling tokens with high classification scores will lead to performance degradation on pose prediction tasks. Therefore, an extra GIoU cost \cite{rezatofighi2019generalized} is calculated when matching the positive samples.  In addition, the network predicts the IoU quality, which is supervised by positive samples generally. However, the usage of one-to-one assignment leads to unfairness, which means that the background often has high confidence prediction. Inspired by Generalized Focal loss \cite{li2020generalized}, we modify the classification branch to jointly estimate the classification-IoU quality $\mathcal{Q}$. In this way, the quality score of both positive and negative samples can be fairly evaluated, which can be formulated as follows:
\begin{equation}
   \mathcal{L}_Q=-|y-\mathcal{Q}|^\beta\ ((1-y)log(1-\mathcal{Q})+\ ylog(\mathcal{Q}))
\end{equation}
where $y$ is the IoU for positive matching pairs while equals 0 for the negatives. The parameter $\beta$ is the modulating factor to down-weight easy examples \cite{lin2017focal} (we set it to 2.0 following common practice).

\textbf{Centroid-aware Sampling.} PETR \cite{liu2022petr} infers 3D attributes by learning the relationship between queries and 3D implicit features. This interaction is equivalent to learn the projection process from queries' reference point to image plane, as shown in Figure \ref{speed_vs_acc}. We conclude that the accurate distillation of the projected 2.5D centers can help the network locate the objects. Therefore, we set an auxiliary task of 2.5D center offset for each location on feature map and supervise it with L1 loss. Besides, a key point prediction network is trained to give a high confidence on features around centroid. The Gaussian heatmap $\mathcal{H}$ is used to define the ground truth of 2.5D centers $c$:
\begin{equation}
    \mathcal{H}=\exp(-\frac{\left(x-c_x\right)^2+\left(y-c_y\right)^2}{2\delta_{\min{\left(l^\ast,t^\ast,r^\ast,b^\ast\right)}}})
\end{equation}
where $c_x$ and $c_y$ are rounding coordinates of the 2.5D center on the feature map. $\delta_{\min{\left(l^\ast,t^\ast,r^\ast,b^\ast\right)}}$ is the size-adaptation coefficient controlled by normalized distances $(l^\ast,t^\ast,r^\ast,b^\ast)$ from the 2.5D center to the four sides of the bounding box. We use the variant of Focal loss proposed in CenterNet \cite{zhou2019objects} to densely supervise the centerness $\mathcal{c}$.

\subsection{Spatial Alignment Module}

In PETR \cite{liu2022petr}, the implicit positon encoding is the only way to distinguish the camera poses, as shown in Equation \ref{eq3}, which causes ambiguity in overlapping areas. In addition, the instance-level down-sampling leads to the loss of global receptive field weakening the depth estimation. In order to align embedding space, the spatial alignment module (see Figure \ref{fig4}) is proposed to transform the 2D features $\boldsymbol{\mathbit{F}}_j^{u,\ v}$ from image plane to a 3D unified space:
\begin{equation}
    \label{spatial}
    {\boldsymbol{\mathbit{F}}^\ast}_j^{u,\ v}=\ \mathcal{T}_w(\boldsymbol{\mathbit{r}}_j^{u,v},{\ \boldsymbol{\mathbit{f}}_j)\ \cdot\boldsymbol{\mathbit{F}}_j^{u,\ v}+\mathcal{T}_b(\boldsymbol{\mathbit{r}}_j^{u,v},{\ \boldsymbol{\mathbit{f}}}_j)}
\end{equation}
where $\mathcal{T}_w$ and $\mathcal{T}_b$ are two feedforward networks to encode intrinsic ${\ \boldsymbol{\mathbit{f}}}_j$  and tracing ray $\boldsymbol{\mathbit{r}}_j^{u,v}$ of a specific camera. 

\subsection{Training and Inference.}  The priority of feature sampling $\mathcal{P}$ depends on both the quality score $\mathcal{Q}$ and the centerness $\mathcal{C}$ during the training or inference stage:
\begin{equation}
	\mathcal{P}=\mathcal{Q}^{\ \alpha}\mathcal{C}^{1-\alpha}
\end{equation}
We use $\alpha$  to balance the weights in the sampling process. Following PnP-DETR \cite{wang2021pnp}, we dynamically select top $\rho$ ratio features in the training phase and set a fixed threshold $\rho^\ast$ for inference. The proposed focal sampling module is trained end-to-end and the auxiliary loss $\mathcal{L}_Q$ is plugged to the original 3D branch\cite{wang2022detr3d, liu2022petr}.:
\begin{equation}
\begin{split}
    \mathcal{L}_Q=\ \frac{1}{N_{pos}}(\lambda_1\mathcal{L}_Q+{\lambda_2\mathcal{L}}_{2.5D}\ {+\ \lambda}_3\mathcal{L}_{GIoU} \\
     \ {+\ \lambda}_4\mathcal{L}_{ltrb} +\ \lambda_5\mathcal{L}_{centerness})
\end{split}
\end{equation}
where we adopt the weighted values ${\ \lambda}_{1-5}$  as 2, 10, 5, 2, 1, respectively.

\begin{table*}[t]
\centering
\caption{Comparison of 3D object detection performance on nuScenes val set. $\ast$ indicates that the models are trained with CBGS\cite{zhu2019class} strategy. $\dagger$ notes using the pre-trained FCOS3D backbone. “S” indicates model with a single time stamp input.}
\vspace{-0.3cm}
\label{val_tab}
\resizebox{\textwidth}{40mm}{
\begin{tabular}{l|c c|c c c c c c c|c}
\toprule
Methods & Backbone & Resolution & mAP$\uparrow$ & NDS$\uparrow$ & mATE$\downarrow$ & mASE$\downarrow$ & mAOE$\downarrow$ & mAVE$\downarrow$ & mAAE$\downarrow$ & FPS$\uparrow$ \\
\midrule
BEVDet$\ast$\cite{huang2021bevdet} & Res-50 & $ 704 \times 256 $ & 0.298 & 0.379 & 0.725 & 0.279 & 0.559 & 0.860 & 0.245 & 16.7 \\
BEVDepth-S\cite{li2022bevdepth} & Res-50 & $ 704 \times 256 $ & 0.315 & 0.367 & \textbf{0.702} & 0.271 & 0.621 & 1.042 & 0.315 & 15.7 \\
PETR$\ast$\cite{liu2022petr} & Res-50 & $ 1408 \times 512 $ & 0.339 & 0.403 & 0.748 & 0.273 & 0.539 & 0.907 & \textbf{0.203} & 8.1 \\
\midrule
Focal-PETR-H-0.33 & Res-50 & $ 704 \times 256 $ & 0.311 & 0.388 & 0.780 & 0.281 & 0.545 & 0.867 & 0.204 & \textbf{30.0} \\
Focal-PETR-H-0.5 & Res-50 & $ 800 \times 320 $ & 0.335 & 0.406 & 0.746 & \textbf{0.270} & \textbf{0.523} & \textbf{0.862} & 0.220 & 22.9 \\
Focal-PETR & Res-50 & $ 800 \times 320 $ & 0.320 & 0.381 & 0.788 & 0.278 & 0.595 & 0.893 & 0.228 & 20.2 \\ 
Focal-PETR & Res-50 & $ 1408 \times 512 $ & \textbf{0.363} & \textbf{0.414} & 0.760 & 0.279 & 0.538 & 0.876 & 0.224 & 8.5 \\ 
\midrule
\midrule
FCOS3D\cite{wang2021fcos3d} & Res-101 & $ 1600 \times 900 $ & 0.295 & 0.372 & 0.806 & 0.268 & 0.511 & 1.315 & \textbf{0.170} & 1.7 \\
PGD\cite{wang2022probabilistic} & Res-101 &  $ 1600 \times 900 $ & 0.335 & 0.409 & 0.732 & 0.263 & 0.423 & 1.285 & 0.172 & 1.4 \\
EPro-PnP\cite{chen2022epro} & Res-101 & $ 1600 \times 900 $ & 0.352 & 0.430 & 0.667 & \textbf{0.258} & \textbf{0.337} & 1.031 & 0.193 & 3.3 \\
DETR3D$\ast$ $\dagger$\cite{wang2022detr3d} & Res-101 & $ 1600 \times 900$  & 0.349 & 0.434 & 0.716 & 0.268 & 0.379 & 0.842 & 0.200 & 3.7 \\
BEVFormer-S$\dagger$\cite{li2022bevformer} & Res-101 & $ 1600 \times 900 $ & 0.375 & 0.448 & 0.725 & 0.272 & 0.391 & \textbf{0.802} & 0.200 & 3.0 \\
Ego3RT$\dagger$\cite{lu2022learning} & Res-101 & $ 1600 \times 900 $ & 0.375 & 0.450 & \textbf{0.657} & 0.268 & 0.391 & 0.850 & 0.206 & 1.7 \\
SpatialDETR$\dagger$\cite{dollspatialdetr} & Res-101 & $ 1600 \times 900 $ & 0.351 & 0.425 & 0.772 & 0.274 & 0.395 & 0.847 & 0.217 & 3.5 \\
PETR$\ast$ $\dagger$\cite{liu2022petr} & Res-101 & $ 1408 \times 512 $ & 0.366 & 0.441 & 0.717 & 0.267 & 0.412 & 0.834 & 0.190 & 5.7 \\
\midrule
Focal-PETR-H-0.5$\dagger$ & Res-101 & $ 1408 \times 512 $ & 0.390 & \textbf{0.461} & 0.678 & 0.263 & 0.395 & 0.804 & 0.202 & \textbf{6.6} \\
Focal-PETR$\dagger$ & Res-101 & $ 1600 \times 640 $ & 0.385 & 0.448 & 0.737 & 0.265 & 0.404 & 0.831 & 0.207 & 4.4 \\
Focal-PETR-H-0.5$\dagger$ & Res-101 & $ 1600 \times 640 $ & \textbf{0.393} & 0.457 & 0.695 & 0.264 & 0.383 & 0.850 & 0.206 & 4.9 \\
\bottomrule
\end{tabular}}
\end{table*}


\section{Experiment}
\label{sec:experiment}

\subsection{Dataset and Metircs}
We validate our proposed Focal-PETR on the large-scale nuScenes dataset \cite{caesar2020nuscenes}, which is the most frequently used dataset for vision-centric perception with 6 calibrated cameras covering a 360-degree horizontal FOV. This dataset consists of 1000 driving scenes, which are officially separated into 700/150/150 scenes for training, validation and testing. Specifically, each scene is of 20s duration and fully annotated every 0.5s. Following common practice, we report official metrics with NuScenes Detection Score (NDS), mean Average Precision (mAP) and 5 kinds of True Positive (TP) metrics including average translation error (ATE), average scale error (ASE), average orientation error (AOE), average velocity error (AVE), average attribute error (AAE).

\subsection{Implementation Details}
To verify the effectiveness of our method under different pre-training, three types of image encoders are employed: ResNet-50, ResNet-101 \cite{he2016deep} and VoVNet-99 \cite{lee2019energy}. Note that, ResNet-50 is initialized from ImageNet \cite{deng2009imagenet} checkpoint, ResNet-101 is initialized from FCOS3D \cite{wang2021fcos3d} checkpoint, and VoVNet-99 is initialized from DD3D \cite{park2021pseudo} checkpoint. The down-sample stride of image encoder is set to 16. We adopt the same image and BEV data augmentation methods as PETR\cite{huang2021bevdet,liu2022petr}. The Transformer \cite{vaswani2017attention} decoder head consists of 6 layers with 900 object queries.

All models are trained using AdamW \cite{loshchilov2017decoupled} optimizer with a total batch size of 16. The learning rate is initialized as 4e-4. There are two variants of our method, Focal-PETR without token sampling and Focal-PETR-H-$\rho^\ast$ with token sampling ratio $\rho^\ast$. For results on nuScenes val set, Focal-PETR and Focal-PETR-H-$\rho^\ast$ are trained for 24 and 60 epochs respectively. For experiments on test set, the training schedules are set up to 100 epochs. Query denoising \cite{li2022dn} is not used in all experiments.

\subsection{State-of-the-art Comparison}
We firstly evaluate Focal-PETR on the nuScenes val set and compare it with the state-of-the-art methods listed in Table \ref{val_tab}, including DETR3D \cite{wang2022detr3d}, BEVDet \cite{huang2021bevdet}, BEVFormer \cite{li2022bevformer}, BEVDepth \cite{li2022bevdepth}, PETR \cite{liu2022petr}, etc. As shown in Table \ref{val_tab}, we set relative small input resolutions with ResNet-50 backbone to compare with light-weight models. Focal-PETR with $704\times256$ resolution input produces impressive inference speed at 30.0 FPS, which is 1.8 times faster than BEVDet. Remarkably, Focal-PETR with size $800\times320$ reaches a desirable trade-off between accuracy and speed. With ResNet-101 backbone, Focal-PETR is trained with $1408\times512$ resolution to fairly compare with PETR. We can see that our method exceeds PETR by 2.4$\%$ in mAP and 2.0$\%$ in NDS though PETR is trained with CBGS strategy. With a larger $ (1600\times640) $ resolution input, Focal-PETR outperforms Ego3RT, the state-of-the-art method with $ (1600\times900) $ resolution input, by 1.8$\%$ in mAP and 0.7$\%$ in NDS respectively.

As shown in Table \ref{test_tab}, we also conduct experiments on nuScenes test set and Focal-PETR produces the outstanding results both on mAP and NDS. On the ResNet-101 backbone, Focal-PETR outperforms BEVFormer, the state-of-the-art method, by 1.7$\%$ in mAP and 2.4$\%$ NDS. It is also noteworthy that with VoVNet-99 \cite{lee2019energy} backbone, Focal-PETR exceeds PETR by 2.4$\%$ in mAP and 1.2$\%$ in NDS.

\begin{figure}[t]
    \centering
    \includegraphics[scale=0.62]{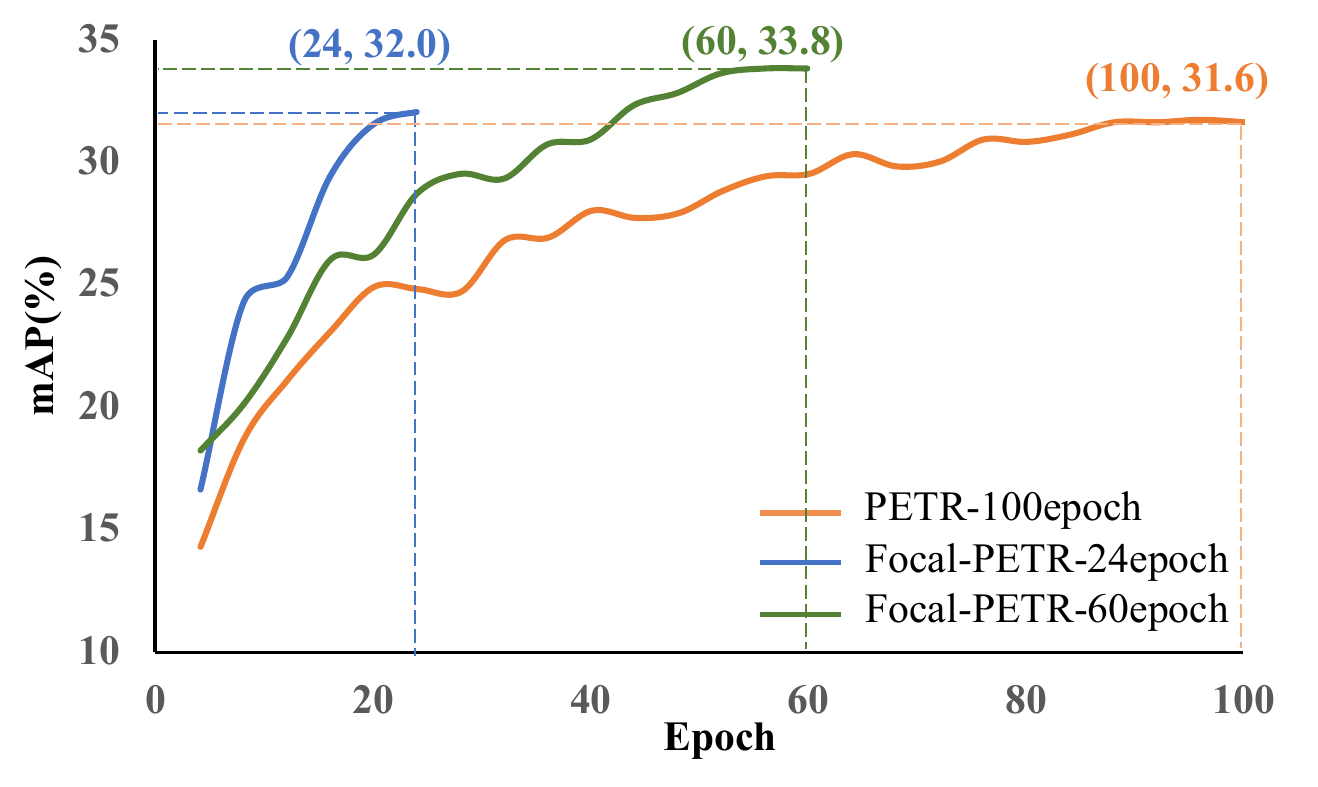}
    \vspace{-0.5cm}
    \caption{Convergence curves of Focal-PETR and PETR. Specifically, we denote the mAP of the last epoch for each curve.}
    \vspace{-0.5cm}
    \label{convergence}
\end{figure}

\begin{table*}[t]
  \centering
  \caption{Comparison of 3D object detection performance on nuScenes test set.}
  \vspace{-0.3cm}
  \label{test_tab}
  \begin{tabular}{l|c|c c c c c c c}
  \toprule
    Methods & Backbone & mAP$\uparrow$ & NDS$\uparrow$ & mATE$\downarrow$ & mASE$\downarrow$ & mAOE$\downarrow$ & mAVE$\downarrow$ & mAAE$\downarrow$ \\
  \midrule
    FCOS3D\cite{wang2021fcos3d} & Res-101 & 0.358 & 0.428 & 0.690 & 0.249 & 0.452 & 1.434 & \textbf{0.124} \\
    EPro-PnP\cite{chen2022epro} & Res-101 & 0.373 & 0.453 & 0.605 & \textbf{0.243} & \textbf{0.359} & 1.067 & \textbf{0.124} \\
    PGD\cite{wang2022probabilistic} & Res-101 & 0.386 & 0.448 & 0.626 & 0.245 & 0.451 & 1.509 & 0.127 \\
    Ego3RT\cite{lu2022learning} & Res-101 & 0.389 & 0.443 & \textbf{0.599} & 0.268 & 0.470 & 1.169 & 0.172 \\
    BEVFormer-S\cite{li2022bevformer} & Res-101 & 0.409 & 0.462 & 0.650 & 0.261 & 0.439 & 0.925 & 0.147 \\
    PETR\cite{liu2022petr} & Res-101 & 0.391 & 0.455 & 0.647 & 0.251 & 0.433 & 0.933 & 0.143\\
  \midrule
    Focal-PETR & Res-101 & \textbf{0.426} & \textbf{0.486} & 0.617 & 0.250 & 0.398 & \textbf{0.862} & 0.146\\
  \midrule
  \midrule
    DD3D\cite{park2021pseudo} & VoV-99 & 0.418 & 0.477 & 0.572 & 0.249 & 0.368 & 1.014 & \textbf{0.124} \\
    DETR3D\cite{wang2022detr3d} & VoV-99 & 0.412 & 0.479 & 0.641 & 0.255 & 0.394 & 0.845 & 0.133 \\
    Ego3RT\cite{lu2022learning} & VoV-99 & 0.425 & 0.473 & 0.549 & 0.264 & 0.433 & 1.014 & 0.145 \\
    BEVDet\cite{huang2021bevdet} & VoV-99 & 0.424 & 0.488 & \textbf{0.524} & \textbf{0.242} & \textbf{0.373} & 0.950 & 0.148 \\
    BEVFormer-S\cite{li2022bevformer} & VoV-99 & 0.435 & 0.495 & 0.589 & 0.254 & 0.402 & 0.842 & 0.131 \\
    SpatialDETR\cite{dollspatialdetr} &  VoV-99 & 0.424 & 0.486 & 0.613 & 0.253 & 0.402 & 0.857 & 0.131 \\
    PETR\cite{liu2022petr} & VoV-99 & 0.441 & 0.504 & 0.593 & 0.249 & 0.383 & \textbf{0.808} & 0.132\\
  \midrule
    Focal-PETR & VoV-99 & \textbf{0.465} & \textbf{0.516} & 0.578 & 0.247 & 0.390 & 0.817 & 0.135\\
  \bottomrule
  \end{tabular}
\end{table*}

\begin{table}[h]
  \centering
  \vspace{-0.2cm}
  \caption{Ablation study of Focal-PETR with different sampling strategies as the sampling ratio is 0.25. Correspondingly, We report the mAP and DNS metrics here. Cls and Ctr denote classification and centerness scores respectively.}
  \vspace{-0.3cm}
  \label{sample_strategy}
  \begin{tabular}{c c c c c|c c}
  \toprule
   & Cls & IoU & Ctr & PnP\cite{wang2021pnp} & mAP$\uparrow$ & NDS$\uparrow$ \\
  \midrule
  (1) & - & - & - & $\checkmark$ & 0.273 & 0.335 \\
  (2) & $\checkmark$ & - & - & - & 0.301 & 0.361 \\
  (3) & $\checkmark$ & $\checkmark$ & - & - & 0.303 & 0.356 \\
  (4) & $\checkmark$ & $\checkmark$ & $\checkmark$ & - & \textbf{0.313} & \textbf{0.372} \\
  
  \bottomrule
  \end{tabular}
\end{table}

\begin{table}[h]
  \centering
  \vspace{-0.2cm}
  \caption{Ablation study with different sampling ratio. Mem. indicates consumption of GPU memory here. To be noted, only computational costs (FLOPs) belonging to detection head are counted. Results with $\ast$ is obtained with vanillia PETR \cite{liu2022petr}.}
  \vspace{-0.3cm}
  \label{sample_ratio}
  \setlength{\tabcolsep}{1.5mm}{
  \begin{tabular}{c|c c|c c c}
  \toprule
  ratio & mAP$\uparrow$ & NDS$\uparrow$ & FLOPs(G) & Mem.(G) & FPS\\
  \midrule
  0.25 & 0.313 & 0.372 & \textbf{24.1(-44.0$\%$)} & \textbf{3.6(-43.8$\%$)} & \textbf{24.0} \\
  0.5 & 0.319 & 0.378 & 30.5(-29.1$\%$) & 4.5(-29.7$\%$) & 22.9 \\
  0.75 & \textbf{0.320} & \textbf{0.379} & 34.8(-14.1$\%$) & 5.5(-14.3$\%$) & 21.2 \\
  1.0 & \textbf{0.320} & \textbf{0.379} & 40.1(+10.2$\%$) & 6.4(+1.6$\%$) & 20.7 \\
  1.0$\ast$ & 0.286 & 0.339 & 36.5 & 6.3 & 20.7 \\
  \bottomrule
  \end{tabular}}
\end{table}

\begin{table}[h]
  \centering
  \vspace{-0.2cm}
  \caption{Different designs of spatial alignment module. We analyze various network designs (module) and feature selection (content). The ray representation only considers the direction of pixel rays, while cone includes the volume viewed by each ray.}
  \vspace{-0.3cm}
  \label{design_spatial}
  \setlength{\tabcolsep}{2mm}{
  \begin{tabular}{c c|c c}
  \toprule
  module & content & mAP$\uparrow$ & NDS$\uparrow$ \\
  \midrule
  - & - & 0.314 & 0.362 \\
  pos. & ray & 0.303 & 0.350\\
  ours. & ray & 0.319 & 0.371 \\
  SE\cite{hu2018squeeze} & cone & 0.316 & 0.380 \\
  ours. & cone & \textbf{0.320} & \textbf{0.381} \\
  \bottomrule
  \end{tabular}}
\end{table}

\begin{table}[h]
  \centering
  \vspace{-0.2cm}
  \caption{Performance Comparison of different sampling ratio on previous and current frame for temporal extension. We additionally report the results of PETRv2\cite{liu2022petrv2} and marked with $\ast$.}
  \vspace{-0.3cm}
  \label{pre_cur_ratio}
  \setlength{\tabcolsep}{2mm}{
  \begin{tabular}{c c|c c c}
  \toprule
  previous & current & mAP$\uparrow$ & NDS$\uparrow$ & mAVE$\downarrow$ \\
  \midrule
  0.2 & 0.5 & 0.337 & 0.437 & 0.443 \\
  0.2 & 1.0 & 0.339 & 0.439 & 0.444 \\
  0.5 & 1.0 & 0.340 & \textbf{0.442} & 0.426 \\
  1.0 & 1.0 & \textbf{0.341} & \textbf{0.442} & \textbf{0.425} \\
  1.0$\ast$ & 1.0$\ast$ & 0.306 & 0.412 & 0.471 \\
  \bottomrule
  \end{tabular}}
\end{table}

For convergence speed, we train Focal-PETR with 24, 60 epochs and PETR with 100 epochs respectively. Both two methods are evaluated every 4 epochs on nuScenes val set visualized in Figure \ref{convergence}. The results show that Focal-PETR achieces higher mAP (32.0\% vs 31.6\%) than PETR even with 3x fewer training epochs. Notably, when comparing the results in Table \ref{val_tab} and Table \ref{test_tab}, Focal-PETR achieves better performance than PETR with different backbone pre-training.

\subsection{Ablation Study}
In this section, we conduct experimental analysis on important components of our methods. All models are trained for 24 epochs without CBGS \cite{zhu2019class} and evaluated on nuScenes val set. More experimental results are presented in the supplementary.

\textbf{Analysis of Focal Sampling.} The results in Table \ref{sample_strategy} show that each component in focal sampling contributes to the performance improvement. Notably, our method with additional instance-guided supervision, outperforms the unsupervised method PnP-DETR, which is inconsistent with the experience in 2D paradigm \cite{wang2021pnp, roh2021sparse}. One possible reason is that the decoder-only architecture is weak in casting potential objects. Under this design, the foreground token is more required. In addition, the centroid-aware sampling brings much more improvement than IoU-aware sampling, which is not consistent with the conclusion drawn in Generalized Focal Loss\cite{li2020generalized}: IoU always performs better than centerness as a measurement of localization quality. This implies that the accurate estimation of the centroid in image plane is crucial for view transformer to learn the implicit 3D-to-2D projection.

We also conduct experiments on the influence of different sampling ratios in Table \ref{sample_ratio}. We firstly compare the results of our method and PETR without sampling (1.0 sampling ratio). It shows that our method achieves 3.4$\%$ and 4.0$\%$ improvement in mAP and NDS. Note that, the extra 3.7 GFLOPs introduced by the proposed sampling module can be discarded when using 1.0 sampling ratio. Further, when conducting experiments with different sampling ratios in Focal-PETR, they have a more significant influence on computation metrics. The results show that with 0.25 sampling ratio, the model reduces the FLOPs and memory consumption by absolutely 16.0 G and 2.8 G compared with 1.0 sampling ratio. That is to say, it only sacrifices 0.7$\%$ mAP and 0.7$\%$ NDS for nearly 43.8$\%$ fewer memory costs and 44.0$\%$ fewer decoder FLOPs.

\begin{figure*}[t]
    \centering
    \includegraphics[scale=0.17]{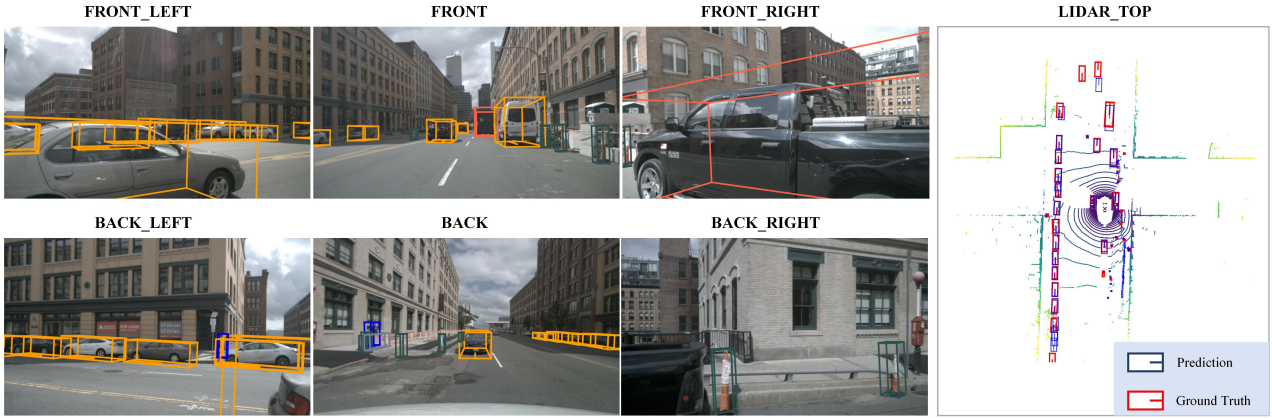}
    \caption{Qualitative detection results of Focal-PETR on nuScenes val set. We show 3D bounding boxes predicted both in multi-camera images and bird's eye view. In multi-camera images, 3D boxes in different colors note different classifications. While in bird's eye view, the ground-truthes are drawn in red to be distinguishable with our predictions in blue.}
    \vspace{-0.3cm}
    \label{visualization}
\end{figure*}

\begin{figure}[t]
    \centering
    \includegraphics[scale=0.55]{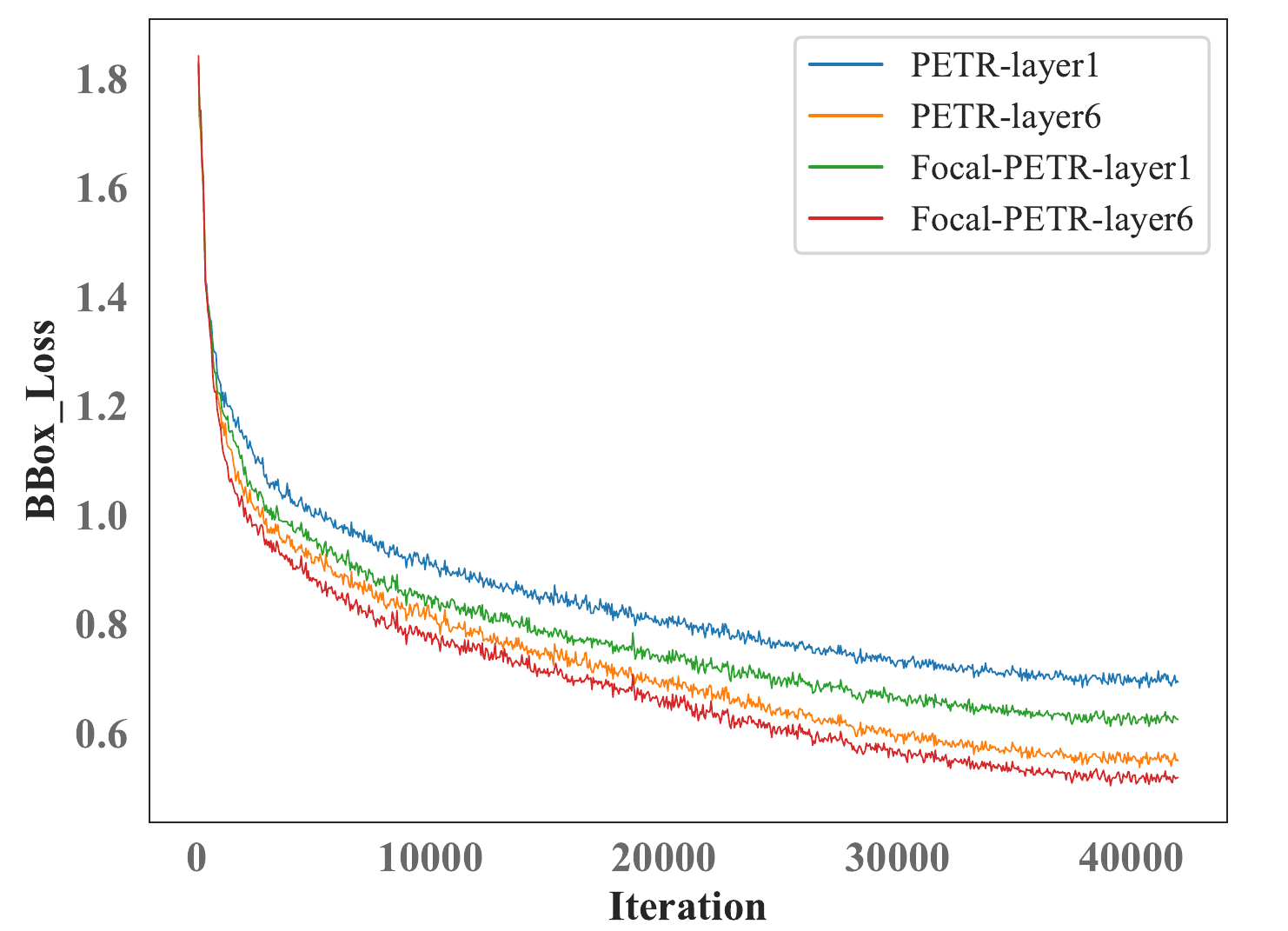}
    \caption{Comparison of L1 losses in the 1st and the 6th layer.}
    \vspace{-0.6cm}
    \label{query_diatance}
\end{figure}

\textbf{Effectiveness of Spatial Alignment.} As shown in Table \ref{design_spatial}, different designs for spatial alignment module lead to large variance in detection accuracy. Simply applying the same position encoding scheme for key-value pairs will impair the performance. It can be seen that our design performs better than the commonly used SE-like architecture \cite{hu2018squeeze}, indicating that the modulation of both weight and bias (see Equation \ref{spatial}) can achieve more flexible spatial transformation. Furthermore, frustum cone provides a more precise geometric prior than pixel ray to perceive the 3D scene, and it brings significantly 0.9\% NDS improvement.

\textbf{Extension in Temporal Modeling.} Existing work \cite{li2022bevdepth, li2022bevformer} has demonstrated that temporal clues can assist network in velocity estiamtion and highly occluded objects detection. To verify the scalability of our method in temporal modeling, we conduct a detail analysis on various sampling ratio of different time stamps. As shown in Table \ref{pre_cur_ratio}, the sampling of past image features have a little impact on the final predictions. Only with 20\% tokens of historical features reaches almost equivalent performance (mAP of 33.9\% vs. 34.1\%, NDS of 43.9\% vs 44.2\%). Remark that, our semantics and spatial alignment strategies still bring great performance improvement. Compared with the PETRv2 baseline, the imporvement in mAP is noticeable of 3.5\%, and the performance gains in NDS is 3.0\%. This phenomenon also implies the potential of our method in temporal modeling.

\subsection{Visualization and Analysis}

We visualize the detection results of Focal-PETR, as illustrated in Figure \ref{visualization}.  We employ Res101 as backbone with $ 1600 \times 640 $ resolution input. The sampling ratio is set as 0.5. Our model is capable of capturing small or dense objects. 

To quantitatively prove that Focal-PETR can efficiently focus on the foreground tokens, we show the L1 loss of bounding boxes predicted by the 1st and the 6th decoder layers in Figure \ref{query_diatance}. Note that, in the 1st decoder layer, the loss of Focal-PETR is significantly less than PETR, indicating that bounding boxes predicted by our method is closer to the objects in the shallow decoder layer.

\section{Conclusion}
We propose Focal-PETR, a multi-camera 3D detection method that mitigates the semantic ambiguity and spatial misalignment of the implicit paradigm. Considering that object detection inherently focuses on foreground information, Focal-PETR takes instance-guided supervision to select discriminative image tokens. These tokens are semantically centralized, which is helpful for the detection head to quickly locate foreground instances. The proposed spatial alignment module enhances the search sensitivity of object query by introducing precise geometric representation. Extensive experiments on the large-scale nuScenes benchmark demonstrate that Focal-PETR achieves state-of-the-art performance and superior efficiency.

Although the proposed Focal-PETR achieves high efficiency in multi-camera 3D object detection, our sampling strategy ignores the representation of map elements, which may hinder the joint learning of object detection and high-quality map segmentation. For future work, we will explore advanced techniques to mitigate this limitation.

\section{Acknowledgement} This work was supported by the National Natural Science Foundation of China (52102449),  the China Postdoctoral Science Foundation (2021M690394), and the Beijing Institute of Technology Research Fund Program for Young Scholars. Besides, we thank Yehansen Chen for helpful discussions.

{\small
\bibliographystyle{ieee_fullname}
\bibliography{egbib}
}

\newpage
\appendix
\section*{Supplementary Material}

\section{Overview}
This document provides more details of network architecture, additional experimental results and qualitative results of ablation study and visualization to the main paper.
\section{Detailed Network Architecture}
\subsection{Focal Sampling}
\begin{figure}[h]
    \centering
    \includegraphics[scale=0.13]{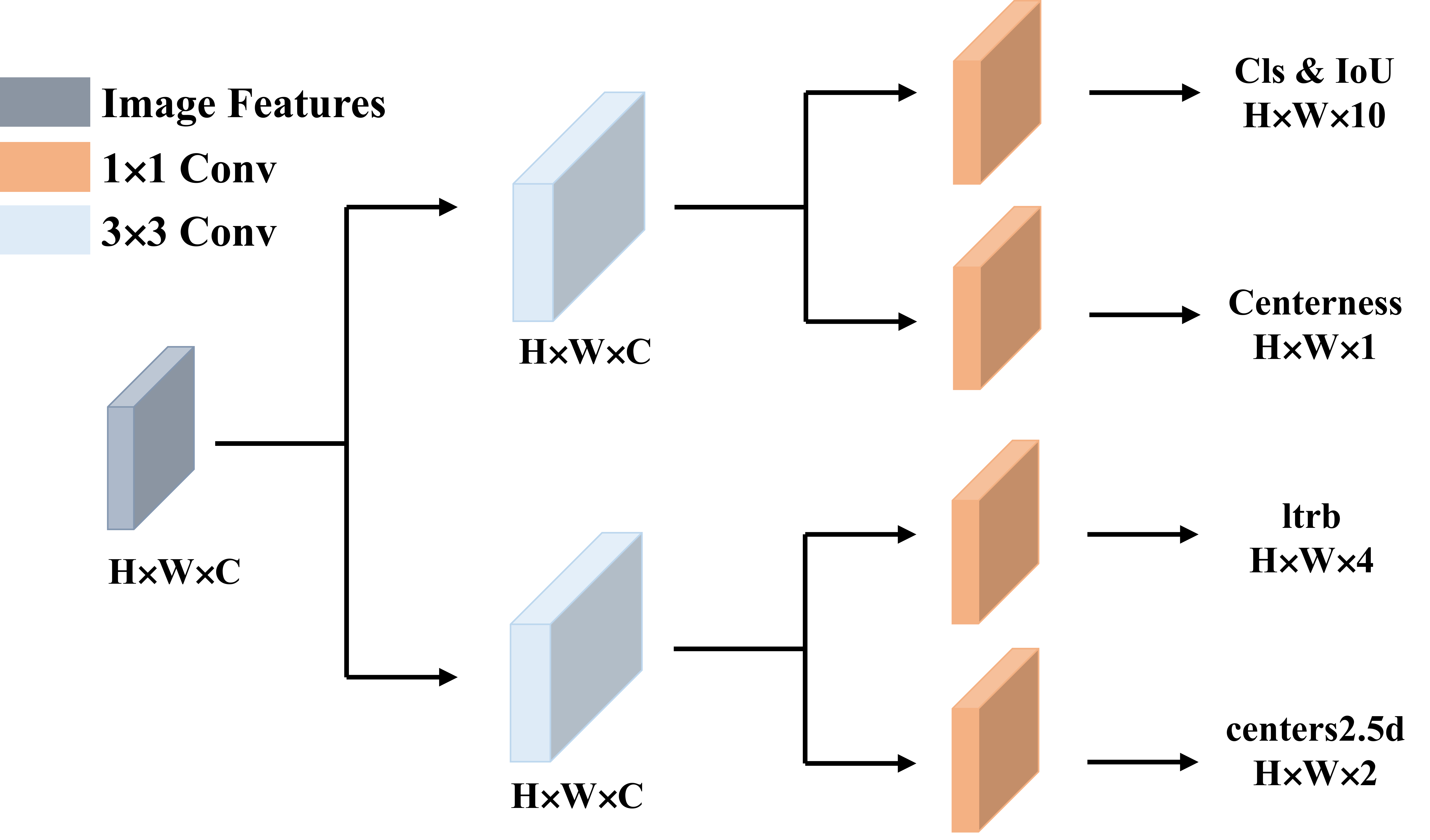}
    \caption{The network design of sampling module. Each image is fed into two light-weight convolution layers to predict the quality scores and attributes. For the supervision of Focal sampling module, please refer to Method in main text. }
    \label{sampling_module}
\end{figure}

\subsection{Spatial Alignment Module}
\begin{figure}[h]
    \centering
    \includegraphics[scale=0.18]{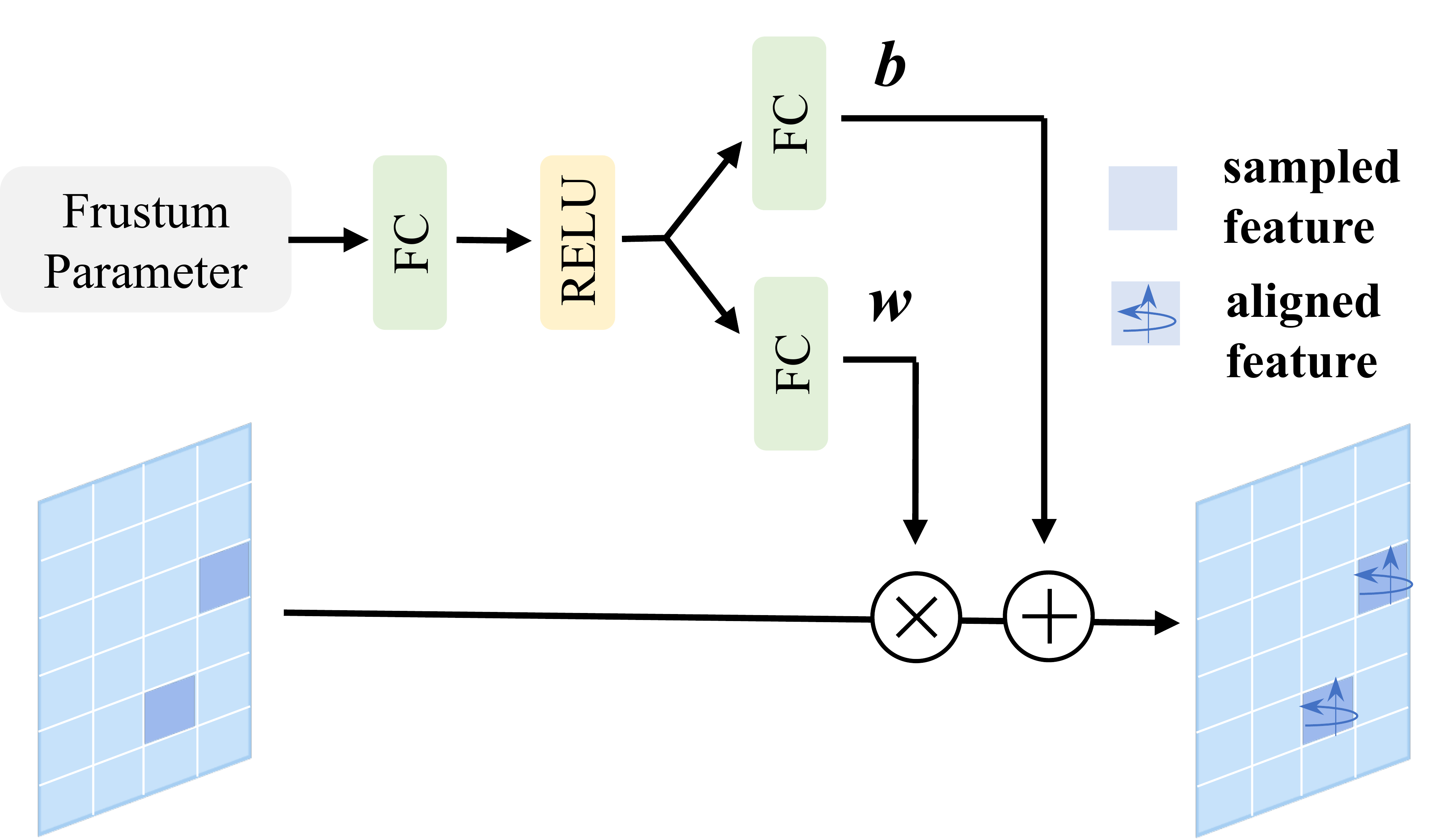}
    \caption{The network design of spatial alignment module. In order to embed spatial information into image features, we use two MLPs to encode parameterized frustum cones. The obtained weight $\boldsymbol{w}$ and bias $\boldsymbol{b}$ are applied to re-weight the image features.}
    \label{spatial_alignment_module}
\end{figure}


\section{Additional Experimental Results}
\begin{figure}[h]
    \centering
    \includegraphics[scale=0.45]{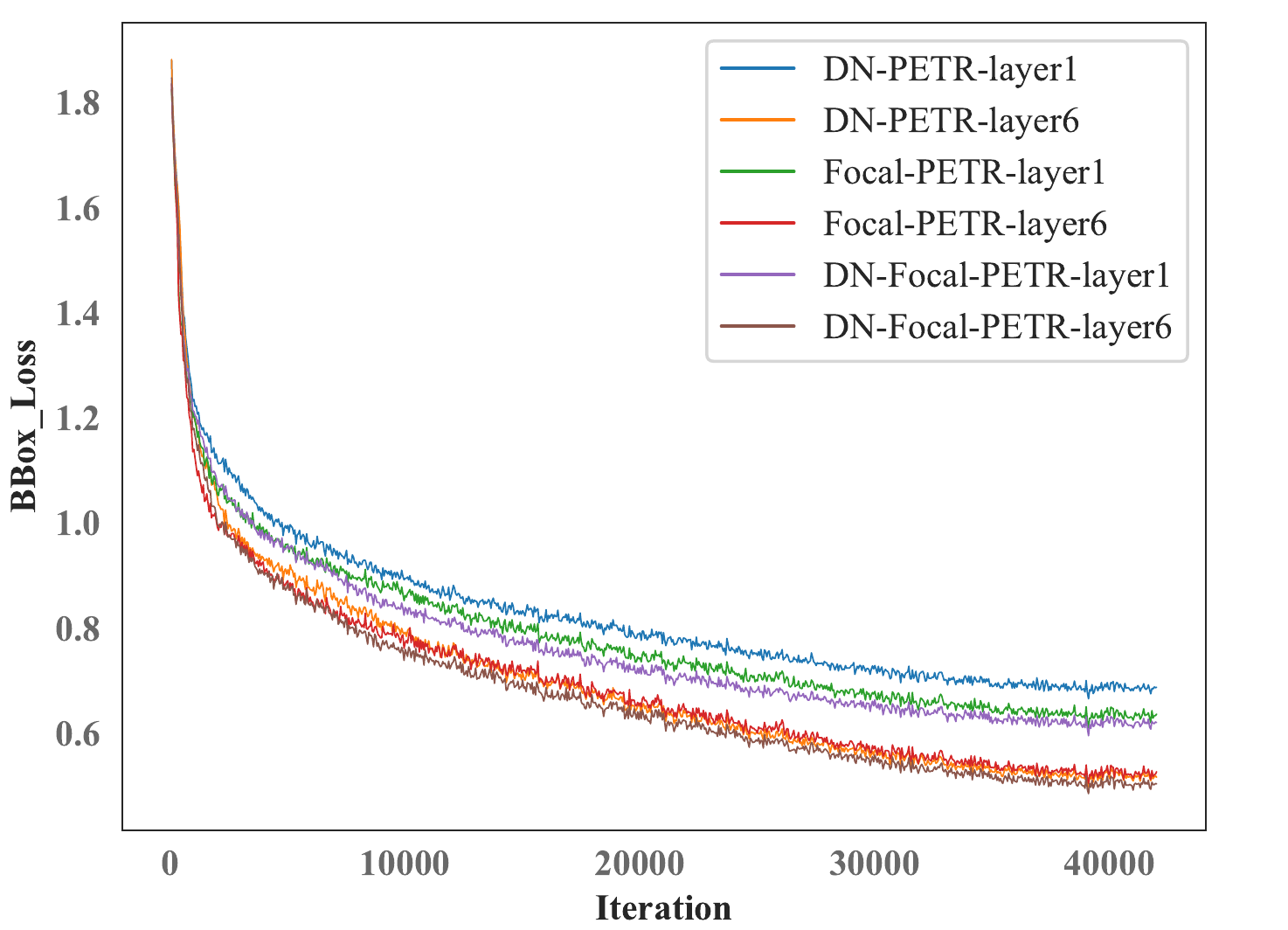}
    \caption{Comparison of regression losses in the 1st and the 6th layers. Note that “DN” indicates models trained with query denoising.}
    \label{bbox_loss}
\end{figure}

\begin{figure}[h]
    \centering
    \includegraphics[scale=0.45]{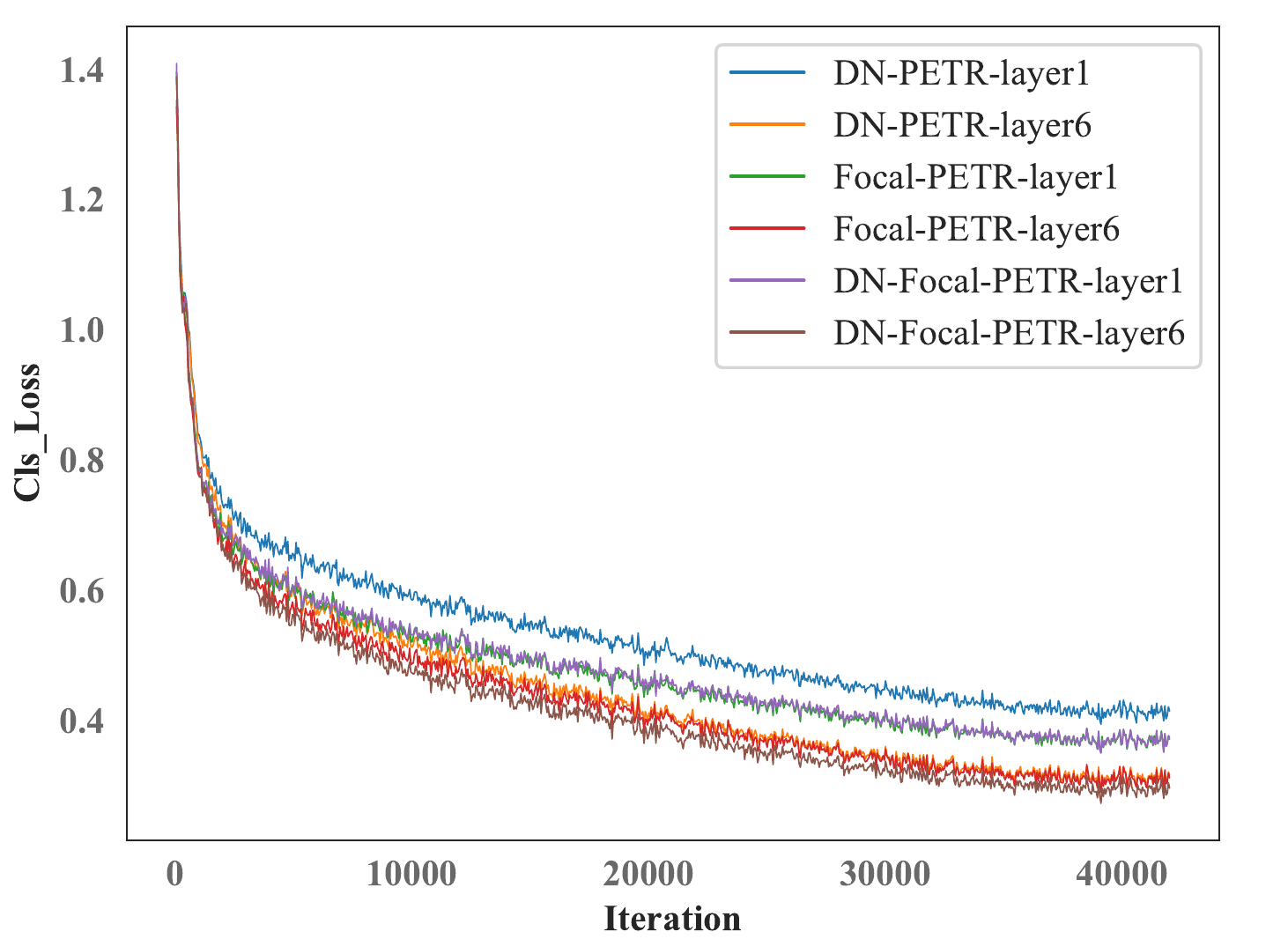}
    \caption{Comparison of classification losses in the 1st and the 6th layers. Note that “DN” indicates models trained with query denoising.}
    \label{cls_loss}
\end{figure}

We train PETR \cite{liu2022petr} and Focal PETR with query denosing \cite{li2022dn} under the same setting. The training loss and evaluation results are shown in Figure \ref{bbox_loss} and Figure \ref{cls_loss}.  It can be seen that our method is superior to query denosing in multi camera 3D detection. Focal PETR can still benefit from query denosing.

More experimental and visualization results are as follows.

\begin{table*}[t]
  \centering
  \caption{Complete results of Focal-PETR with different sampling strategies as the sampling ratio is 0.25. Cls and Ctr denote classification and centerness scores respectively. We also provide the evaluation results of the model only using C5 feature without sampling. It has worse performance and comparable computational cost compared with the models using P4 feature with 0.25 sampling ratio.}
  \label{sample_strategy}
  \begin{tabular}{c c c c c|c c|c c c c c}
  \toprule
   & Cls & IoU & Ctr & feature & mAP$\uparrow$ & NDS$\uparrow$ & mATE$\downarrow$ & mASE$\downarrow$ & mAOE$\downarrow$ & mAVE$\downarrow$ & mAAE$\downarrow$ \\
  \midrule
  (1) & $\checkmark$ & - & - & P4 & 0.301 & 0.361 & 0.831 & 0.279 & 0.634 & \textbf{0.908} & 0.239 \\
  (2) & $\checkmark$ & $\checkmark$ & - & P4 & 0.303 & 0.356 & 0.833 & 0.279 & 0.640 & 0.965 & 0.239 \\
  (3) & $\checkmark$ & $\checkmark$ & $\checkmark$ & P4 & \textbf{0.313} & \textbf{0.372} & \textbf{0.807} & \textbf{0.277} & 0.584 & 0.943 & \textbf{0.236} \\
  (4) & $\checkmark$ & $\checkmark$ & $\checkmark$ & C5 & 0.308 & 0.371 & 0.808 & 0.280 & \textbf{0.576} & 0.911 & 0.247\\
  \bottomrule
  \end{tabular}
\end{table*}

\begin{table*}[t]
  \centering
  \vspace{-0.3cm}
  \caption{Complete results of different designs for spatial alignment module. We analyze various network designs (module) and feature selection (content). The ray representation only considers the direction of pixel rays, while cone additionally includes the volume viewed by each ray.}
  \vspace{-0.3cm}
  \label{design_spatial}
  \setlength{\tabcolsep}{2mm}{
  \begin{tabular}{c c|c c|c c c c c}
  \toprule
  module & content & mAP$\uparrow$ & NDS$\uparrow$ & mATE$\downarrow$ & mASE$\downarrow$ & mAOE$\downarrow$ & mAVE$\downarrow$ & mAAE$\downarrow$ \\
  \midrule
  - & - & 0.314 & 0.362 & 0.808 & 0.281 & 0.601 & 1.049 & 0.262 \\
  pos. & ray & 0.303 & 0.350 & 0.842 & 0.281 & 0.686 & 0.957 & 0.245 \\
  ours. & ray & 0.319 & 0.371 & 0.800 & 0.278 & 0.587 & 0.984 & 0.241 \\
  SE\cite{hu2018squeeze} & cone & 0.316 & 0.380 & 0.803 & 0.278 & \textbf{0.575} & \textbf{0.885} & 0.242 \\
  ours. & cone & \textbf{0.320} & \textbf{0.381} & \textbf{0.791} & \textbf{0.276} & 0.607 & \textbf{0.885} & \textbf{0.232}\\
  \bottomrule
  \end{tabular}}
\end{table*}

\begin{table*}[t]
  \centering
  \vspace{-0.3cm}
  \caption{Influence of query denoising \cite{li2022dn} on Focal-PETR. We compare the results of Focal-PETR and PETR with query denoising under the same setting.}
  \label{design_spatial}
  \vspace{-0.3cm}
  \setlength{\tabcolsep}{2mm}{
  \begin{tabular}{c c|c c|c c c c c}
  \toprule
  Focal & Denosing & mAP$\uparrow$ & NDS$\uparrow$ & mATE$\downarrow$ & mASE$\downarrow$ & mAOE$\downarrow$ & mAVE$\downarrow$ & mAAE$\downarrow$ \\
  \midrule
  $\checkmark$ & - & 0.321 & 0.381 & 0.800 & 0.280 & \textbf{0.589} & \textbf{0.895} & \textbf{0.235} \\
  - & $\checkmark$ & 0.307 & 0.355 & 0.810 & 0.278 & 0.716 & 0.944 & 0.240 \\
$\checkmark$. & $\checkmark$ & \textbf{0.328} & \textbf{0.382} & \textbf{0.779} & \textbf{0.275} & 0.631 & \textbf{0.895} & 0.240 \\
  \bottomrule
  \end{tabular}}
\end{table*}

\begin{figure*}[t]
\centering
\includegraphics[scale=0.45]{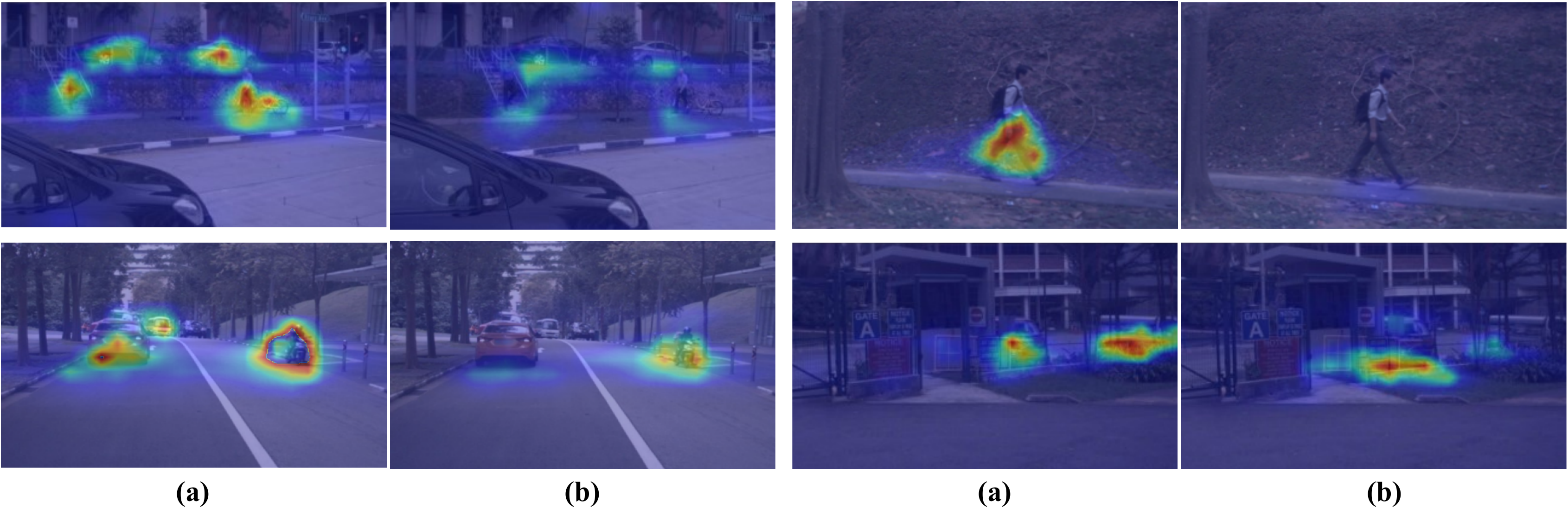}
\vspace{-0.3cm}
\caption{Comparison of attention weight maps for (a) Focal-PETR and (b) PETR with 24 training epochs without CBGS \cite{zhu2019class}. It can be seen that the attention map of Focal-PETR focuses more on the foreground objects.}
\label{attention}
\end{figure*}

\begin{figure*}[t]
    \centering
    \includegraphics[scale=0.18]{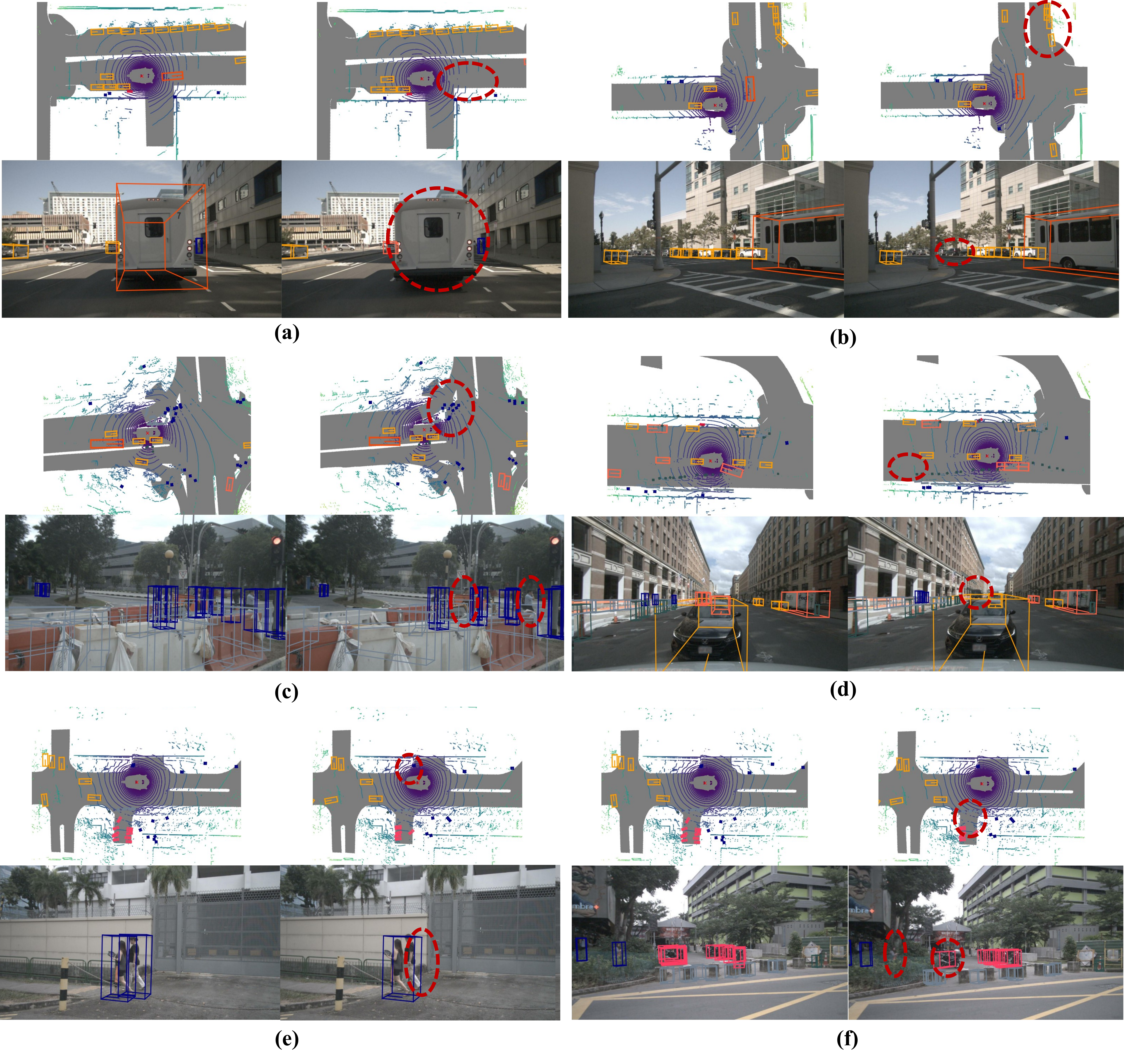}
    \vspace{-0.3cm}
    \caption{Qualitative detection results of Focal-PETR and PETR \cite{liu2022petr} on nuScenes \cite{caesar2020nuscenes} val set. We visualize one of six camera views in each scene and compare the results of two methods. The 3D bounding boxes are projected into BEV and image view. Boxes in different colors note different classifications. In particular,  We note the sub-optimal predictions of PETR in red circles.}
    \vspace{0.1cm}
    \label{visualization}
\end{figure*}

\begin{figure*}[t]
    \centering
    \includegraphics[scale=0.18]{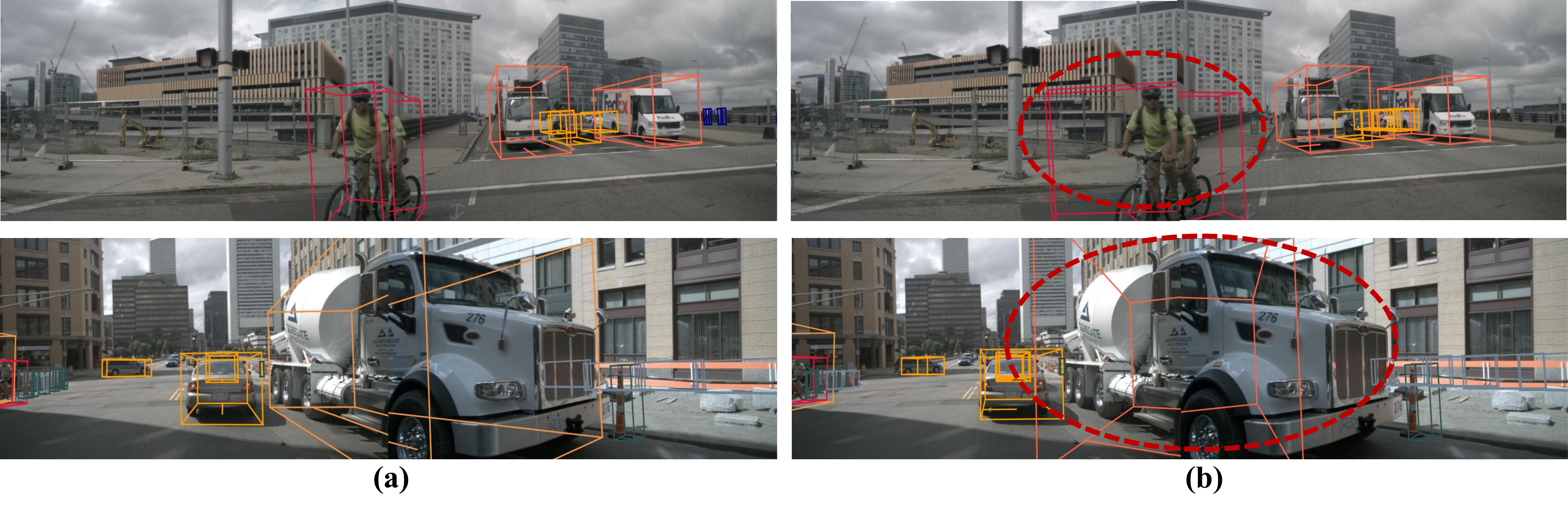}
    \vspace{-0.3cm}
    \caption{Qualitative detection results of (a) Focal-PETR and (b) PETR on nuScenes val set. We visualize multi-camera views in each scene and compare the results of two methods. Boxes in different colors note different classifications. In particular,  we note the sub-optimal predictions of PETR in red circles.}
    \label{visualization}
\end{figure*}

\begin{figure*}[t]
    \centering
    \includegraphics[scale=0.5]{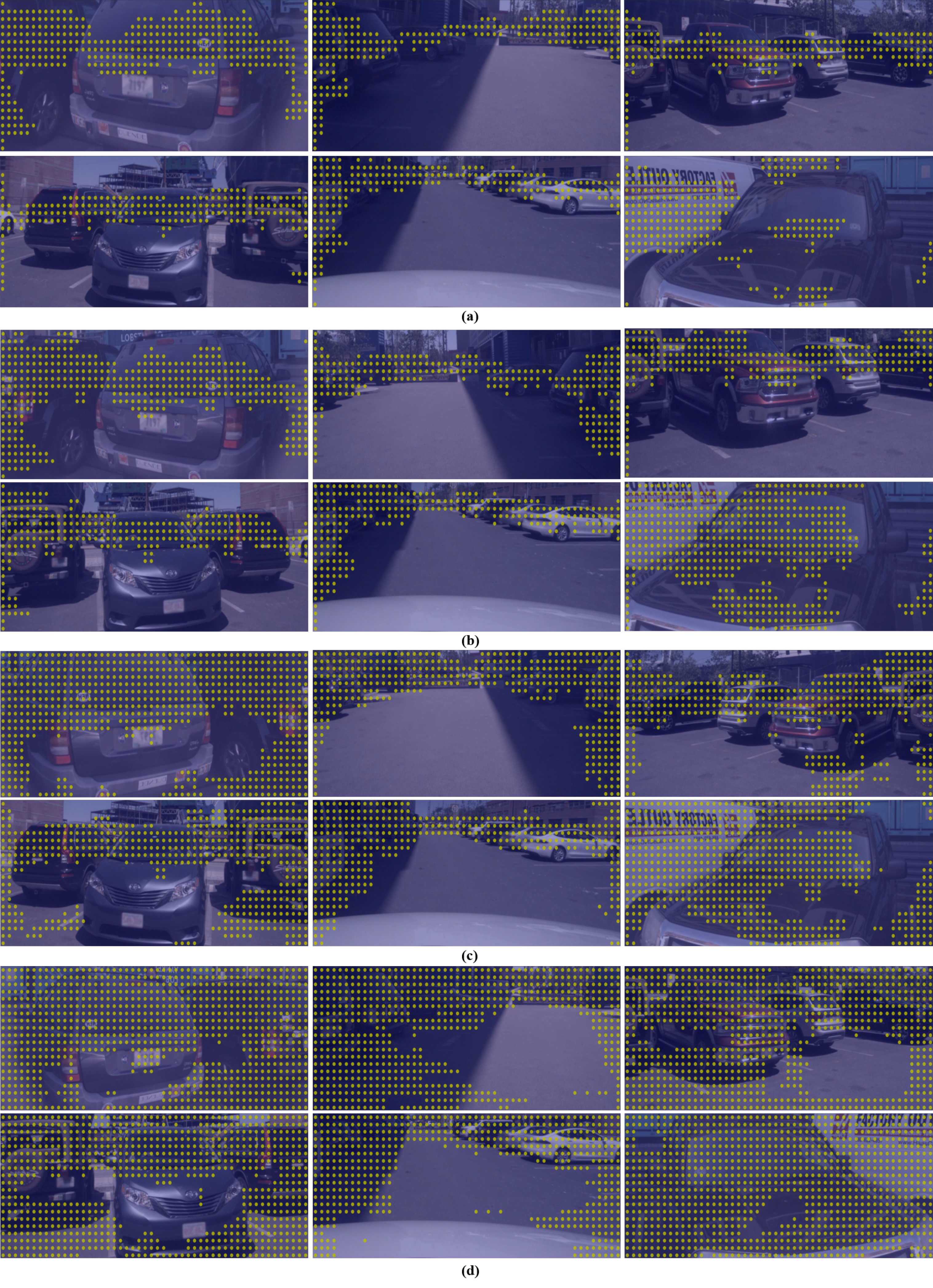}
    \vspace{-0.3cm}
    \caption{Visualization of sampling locations in multi-camera views with different sampling ratios, including (a) 0.25 sampling ratio; (b) 0.33 sampling raito; (c) 0.5 sampling ratio; (d) 0.75 sampling raito. The yellow dots indicate the sampled locations.}
    \vspace{-0.3cm}
    \label{sampling_visualization}
\end{figure*}

\end{document}